\documentclass{article}

\PassOptionsToPackage{round}{natbib}
 \usepackage[preprint]{neurips_2025}

\usepackage[utf8]{inputenc} 
\usepackage[T1]{fontenc}    
\usepackage{hyperref}       
\usepackage{url}            
\usepackage{booktabs}       
\usepackage{amsfonts}       
\usepackage{nicefrac}       
\usepackage{microtype}      
\usepackage{xcolor}         
\usepackage{caption}
\usepackage{svg}
\usepackage{float}

\title{Agentomics-ML: Autonomous Machine Learning Experimentation Agent for Genomic and Transcriptomic Data}

\author{
  Vlastimil Martinek$^{1,2}$ \And
  Andrea Gariboldi$^{1,2}$ \And
  Dimosthenis Tzimotoudis$^{1,2}$ \And
  Aitor Alberdi Escudero$^{1,2}$ \And
  Edward Blake$^{1,2}$ \And
  David \v{C}ech\'{a}k$^{3,4}$ \And
  Luke Cassar$^{1,2}$ \And
    Alessandro Balestrucci$^{1,2}$ \hspace{1.5cm} Panagiotis Alexiou$^{1,2}$ \\
  \\
  $^1$Centre for Molecular Medicine and Biobanking, University of Malta, Msida, Malta\\
  $^2$Department of Applied Biomedical Science, Faculty of Health Sciences, \\ University of Malta, Msida, Malta\\
  $^3$Central European Institute of Technology, Masaryk University, Brno, Czech Republic\\
  $^4$National Centre for Biomolecular Research, Faculty of Science,\\ Masaryk University, Brno, Czech Republic
}

\begin{document}

\maketitle

\begin{abstract}
The adoption of machine learning (ML) and deep learning methods has revolutionized molecular medicine by driving breakthroughs in genomics, transcriptomics, drug discovery, and biological systems modeling. The increasing quantity, multimodality, and heterogeneity of biological datasets demand automated methods that can produce generalizable predictive models. Recent developments in large language model-based agents have shown promise for automating end-to-end ML experimentation on structured benchmarks. However, when applied to heterogeneous computational biology datasets, these methods struggle with generalization and success rates. Here, we introduce Agentomics-ML, a fully autonomous agent-based system designed to produce a classification model and the necessary files for reproducible training and inference. Our method follows predefined steps of an ML experimentation process, repeatedly interacting with the file system through Bash to complete individual steps. Once an ML model is produced, training and validation metrics provide scalar feedback to a reflection step to identify issues such as overfitting. This step then creates verbal feedback for future iterations, suggesting adjustments to steps such as data representation, model architecture, and hyperparameter choices. We have evaluated Agentomics-ML on several established genomic and transcriptomic benchmark datasets and show that it outperforms existing state-of-the-art agent-based methods in both generalization and success rates. While state-of-the-art models built by domain experts still lead in absolute performance on the majority of the computational biology datasets used in this work, Agentomics-ML narrows the gap for fully autonomous systems and achieves state-of-the-art performance on one of the used benchmark datasets. The code is available at \url{https://github.com/BioGeMT/Agentomics-ML}.
\end{abstract}

\section{Introduction} \label{sec:intro}
High throughput -omics technologies, including genomics, transcriptomics, proteomics, metabolomics and epigenomics, have transformed molecular biology into a data driven science. A single high-throughput DNA/RNA sequencing run can generate gigabases of nucleotide data, while each single mass spectrometry experiment quantifies tens of thousands of proteins in parallel. Beyond pure volume, the complexity of molecular biology datasets makes them difficult to model and interpret \citep{Bharti2024}. Despite these issues, research on large molecular biology datasets is crucial for tasks ranging from sequence classification and annotation at the fundamental biology level, up to translational tasks in biomedical science such as disease prediction. Without algorithmic frameworks that can perform quality control, normalization, feature extraction, and other downstream tasks such as classification, these datasets remain collections of uninterpreted measurements rather than sources of biological insight \citep{Xu2019}.

Genomic data production is increasing at an exponential rate, with sequencing throughput doubling approximately every seven months. These datasets are characterized by high dimensionality, complex missing value patterns and technical noise. Batch effects from different instrument runs or library preparations can obscure true biological signals \citep{Yu2024}. Class imbalance is common in tasks such as regulatory element detection or rare variant analysis, in which relevant events may represent less than one percent of the data \citep{Schubach2017}. Combining heterogeneous data modalities such as sequence, expression and metabolite profiles further complicates analysis. As sequencing and detection costs decline the number of skilled analysts has increased only modestly, creating a bottleneck in the translation of raw data into actionable knowledge. The need for further automation of analytical steps is pressing \citep{Wrheide2021}.

In this study we present Agentomics-ML, an agentic system designed to function as a virtual machine learning expert for -omics data analysis. We demonstrate the use of Agentomics-ML on several classification tasks on genomic and transcriptomic data. The system can plan and self-reflect, allowing for decomposition of complex tasks into subtasks that it can iteratively complete and evaluate. We demonstrate that our system can successfully produce working ML code including trained models over 93\% of the time, even in a more complex dataset where all other agentic state of the art systems tested fail to produce working code. Additionally, we offer a lightweight implementation that allows easy deployment and extension by bioinformaticians that may not be ML experts. Using minimal tool use, Agentomics-ML can perform end to end analysis without human intervention, and outperform both zero-shot LLM code, and all benchmarked state of the art agentic ML systems. The code in our experiments is provided in the supplementary materials under the MIT License.

\section{Related work} \label{sec: rel_work}
\subsection{Auto-ML}
Traditional Automated Machine Learning (AutoML) systems, such as AutoSklearn \citep{autoskl}, represent an established approach to automating ML workflows. These systems typically operate by performing automated feature engineering, hyperparameter optimization, and model ensembling, primarily demonstrating strong performance on well-structured tabular data. However, their operational model assumes that input data are already adequately represented and they possess limited inherent capabilities to incorporate domain-specific knowledge or complex user requirements. A key limitation is their general inability to design novel or highly complex model architectures, such as those required for deep learning applications \citep{azevedo2024multivo}, which restricts their utility for challenging biological datasets where custom architectures are often essential for achieving sufficient performance. While these tools require minimal code, a human is still necessary to write code and provide preprocessed data \citep{Sun_2023}.

\subsection{Zero-shot LLMs}
Large Language Models (LLMs) have enabled an alternative modality for ML task automation through their capacity for zero-shot code generation. In this approach, the LLM is provided with a task specification via a prompt, and it generates a code block intended for external execution to train an ML model. This method offers substantial flexibility and can integrate domain knowledge or user requirements expressed in the prompt. However, such systems fundamentally lack the intrinsic ability to directly execute or validate the generated code, nor can they autonomously refine their proposed solutions based on runtime feedback or errors. Consequently, human oversight and intervention are typically required to ensure the correctness and successful execution of the generated code.

\subsection{Agents}
LLM Agents extend LLMs by providing them with tools such as code execution or retrieval. By interacting with the environment through tools they react to errors and intermediate observations, thereby enabling a more iterative and autonomous ML development process \citep{sun2024surveylarge}. Some agentic frameworks are designed for human-in-the-loop operation, where user guidance directs the agent's actions. Others, like DS-Agent, aim for end-to-end automation but may depend on a curated database of solutions to similar past tasks for optimal performance. For instance, SELA \citep{hong2024metagpt} conceptualizes ML pipeline development as a search problem, employing a Monte Carlo Tree Search guided by an LLM to explore and refine data preprocessing, feature engineering, model selection, and hyperparameter configurations, using validation metrics as feedback. Data Interpreter constructs a hierarchical task graph to decompose complex analytical problems, dynamically invoking external tools for sub-tasks like model training and statistical analysis, and incorporates mechanisms for consistency checking and backtracking. AIDE \citep{jiang2025aideaidri} adopts a tree-search methodology, treating each potential ML solution as a node and iteratively refining promising candidates using tools for file management and script evaluation, achieving notable results on benchmarks including MLE-Bench.

\subsection{Benchmarks}
Despite these advancements, current agentic ML systems commonly face significant challenges in evaluation and deployment. Evaluations are frequently confined to Kaggle-style tabular benchmarks, which may not adequately represent the complexities of domain-specific -omics datasets. While initiatives like MLAgentBench \citep{huang2024mlagent} and MLE-Bench \citep{chan2025mlebench} are expanding evaluation to diverse tasks, the extensive prior exploration of some benchmark datasets may influence LLM decisions if their training data includes existing solutions. A critical issue is the robust prevention of test set information leakage; many systems lack strict programmatic abstraction of the test data from the LLM during the development phase, potentially inflating reported performance metrics. Furthermore, existing solutions often do not produce fully reproducible training environments or production-ready inference scripts without considerable manual post-processing. These gaps underscore the need for domain-aware agentic frameworks. To this end, curated resources like Genomic Benchmarks \citep{Greov2023} provide standardized datasets for genomic sequence classification, and miRBench \citep{Sammut2024} offers novel, bias-corrected datasets for microRNA binding site prediction, facilitating rigorous and fair benchmarking critical for advancing ML in specialized biological domains

\section{Methodology} \label{sec:methodology}
Agentomics-ML is an agentic system constructed to automate machine learning experimentation for -omics data, balancing a structured high-level workflow with significant agent autonomy to produce a suitable machine learning model. Requiring only training data file and optional data description, the system outputs an inference script including all necessary artifacts, enabling immediate inference on external data like a test set. Additionally, our system outputs a training script that contains the training code, ensuring reproducibility.

\subsection{Pre-defined steps}
\begin{figure}
    \centering
    \includegraphics[width=0.8\linewidth]{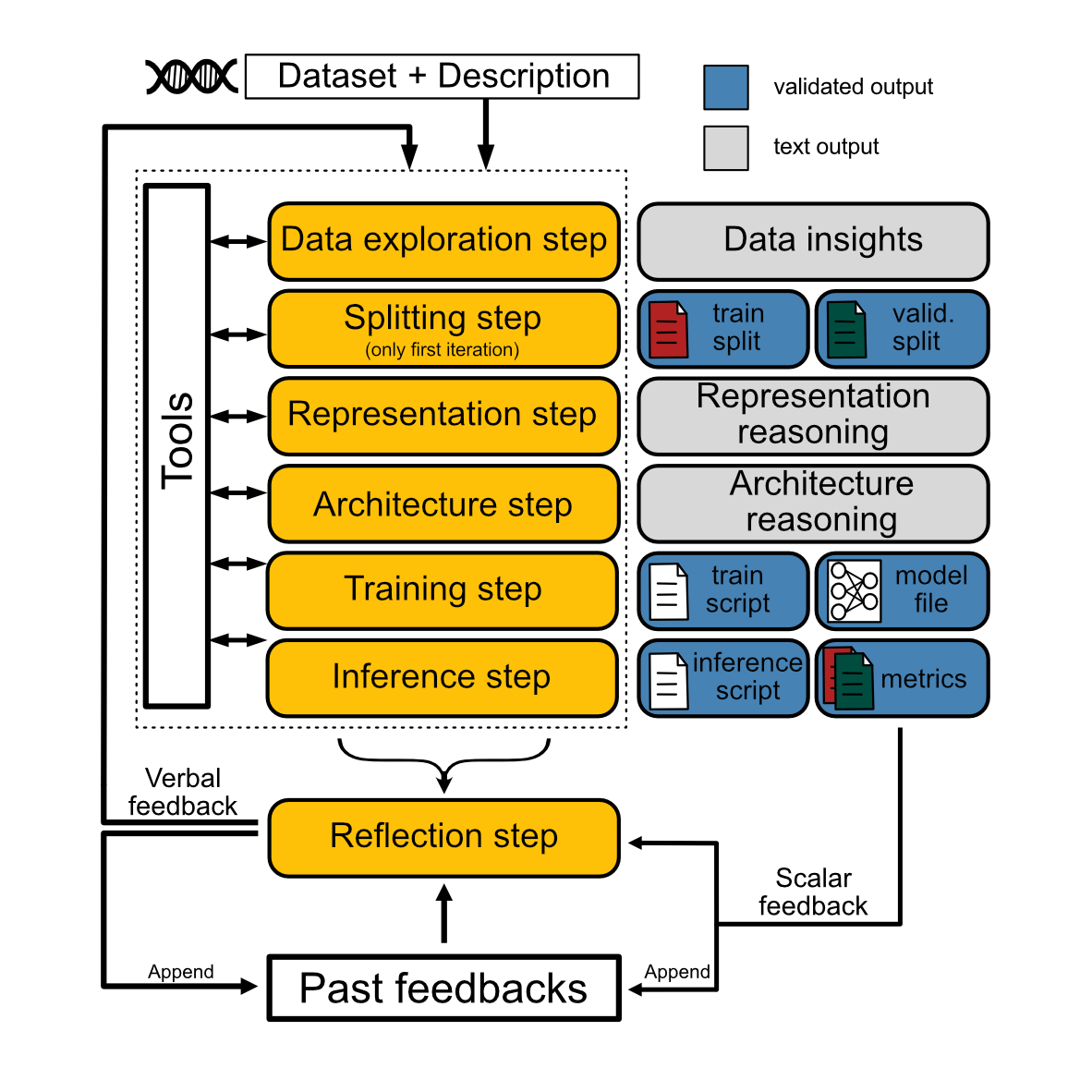}
    \caption{Architecture of Agentomics-ML: Agent follows pre-defined steps of an ML development pipeline, gradually completing sequential steps to eventually output a working ML model. File output steps are validated for correct syntax and format, forcing the agent to retry failed steps. After each iteration both scalar and verbal feedback is collected to guide decision during future iterations.}
    \label{fig:method}
\end{figure}

Agentomics-ML follows pre-defined steps of a machine learning experimentation process as shown in Figure \ref{fig:method}: data exploration, choosing data representation and model architecture, producing training and inference scripts, and training a model. After the first data exploration step, the agent is able to use acquired knowledge like descriptive statistics and domain information from data description to decide on non-random data splitting strategy to better estimate generalization during model development \citep{catania2022random}. We opted for a step-based approach instead of elaborate prompt engineering to avoid unnecessary agent planning \citep{xie2024revealing} as we consider these high-level steps essential in the machine learning experimentation process.

\subsection{Validation of output}
To complete a step, our agent repeatedly uses available tools for bash commands execution, writing python files, and executing python files. Taking actions like installing conda packages and writing helper scripts, the agent eventually outputs either text describing design decisions, acquired knowledge to be used by future steps, or files that are programmatically validated using the Pydantic AI framework \footnote{\url{https://ai.pydantic.dev/}}. For example, when outputting an inference script, the script is run with dummy data to immediately validate the required script arguments and output format of the produced predictions. The agent is triggered to retry and is not able to advance to the next step unless the output passes this programmatic validation. Due to the security risk of providing our agent with a general bash tool and code execution ability, all agent operations occur within a secure, isolated Docker container.

\subsection{Feedback mechanism}
Our agent can run in multiple iterations. Due to the effectiveness for error-correcting capability in complex tasks \citep{huang2024understand, shinn2023reflexion}, we use a reflection step. Processing the whole iteration context and scalar feedback from validation and train set metrics, the agent takes a reflection step producing verbal feedback that is used to guide decisions of the next iteration agent. Verbal feedback is stored together with metrics and additionally used by reflection steps during next iterations. This iterative process allows the system to reflect on outcomes, identify issues such as overfitting by observing validation performance, and progressively improve the model. 

\subsection{Evaluation}
At the very end of this multi-iteration process, we fall back to the model version that achieved the best performance on the validation metric. To prevent data leakage, the agentic system is programmatically abstracted from the held-out test set throughout its development cycle. Only the final, agent-produced inference script is used to report performance on the test set. When comparing Agentomics-ML against other methods, including existing agentic systems and zero-shot LLM approaches, we require that all tested methods are fully automatable and adhere to the same strict test set abstraction to ensure fair and leakage-free evaluation.

\subsection{Datasets}
For benchmarking various methods, we prioritized tasks with existing published state-of-the-art models for comparison \citep{qiao2024model, yu2024selfd, thoutam2024msamamba}. We selected classification datasets from Genomic Benchmarks \citep{Greov2023} that contain genomic sequences for various tasks including detecting enhancers and promoters. Some datasets contain variable-length sequences, a common trait of bioinformatic sequence datasets. Further, we select a classification dataset from miRBench \citep{Sammut2024} that requires modeling of the interaction of two RNA sequences with variable lengths.

\section{Experiments}
\subsection{Success rates}
\subsubsection{LLM zero-shots}
The first baseline of attempting to classify genomic datasets using a non-agentic system is direct use of widely available LLMs with zero-shot prompts. We have selected to evaluate 8 of the most widely used LLMs: OpenAI’s gpt-4.1 and o4-mini, Anthropic’s claude-3.7-sonnet, Google’s gemini-2.5-pro-preview, Alibaba’s qwen3-32b, Deepseek’s deepseek-r1 and deepseek-chat, and Meta’s llama-4-maverick. All models are run with the same prompt (link to supplement) which includes basic data format description and instructions to produce a complete executable solution in a single interaction. The returned code is programmatically parsed, saved, and executed in a controlled environment, without any tool use or environmental interaction by the model. Each model was prompted 5 times, and runs evaluated independently. Success rates for these runs are aggregated per model in Table \ref{tab:max-metrics-sota}. As expected, standalone LLMs fail to consistently produce viable code. Best of class is openAI/o4-mini with a 56.66\% success rate over all datasets. On the other end, qwen/qwen3-32b barely manages to produce working code with a success rate of 6.66\% over all datasets. 

\subsubsection{Agentic systems}
An important added value of agentic systems is their ability to execute code and correct their trajectories based on feedback from the environment, increasing the rate of producing properly running inference scripts. We proceed with benchmarking two state of the art agentic systems that have been developed for producing machine learning models with inference code on arbitrary datasets. Both systems are tested using their publicly available implementations with default settings across all datasets. To ensure comparability, all methods were provided with the standard user prompt, with minimal adjustments to increase method success rates. Similarly to the zero-shot LLM experiments, each agentic system was run 5 times on each dataset, and success rates were calculated and aggregated as before (Table \ref{tab:max-metrics-sota}).
The Data Interpreter (DI) agentic system \citep{hong2024datain} was run with all used LLM models as a backbone. Surprisingly, the success rate of the LLMs did not always improve when used through DI. For the deepseek family of models, DI brought the success rate down to almost zero. On the other hand, OpenAI’s gpt-4.1 model has shown performance improvement.

The second agentic system tested was AIDE \citep{jiang2025aideaidri}. AIDE was configured to use the standard user prompt and was run using gpt-4.1 as the coding model. The success rate was 10\% over all datasets, with 4/6 datasets producing no working inference code in any of the 5 runs. This was primarily due to AIDE scripts containing programming errors, and not producing necessary scripts or artifacts for inference. Importantly, none of the zero-shot LLMs or the agentic systems benchmarked managed to produce any workable code for the most complex of used datasets AGO2\_CLASH\_Hejret. This dataset consists of two sequences per sample, one of which having variable length. Each of these sequences independently has no predictive value, but their interaction is biologically important. All benchmarked methods have 0\% success rate on this dataset, showcasing the failure of such models as soon as moderately complex datasets are introduced.

\subsubsection{Agentomics-ML}
We proceeded with benchmarking our proposed agentic system on the same datasets. Again, the system is run 5 times per dataset, using the same prompt instructions as before, on the same hardware. We notice a significant improvement in success rates, with 100\% success rate for five out of six datasets, including the more complex AGO2\_CLASH\_Hejret dataset. On the sixth dataset we achieve a 60\% success rate, bringing the overall success rate over all datasets to 93.33\%. These results showcase that our solution can consistently produce workable machine learning code and models, even on datasets with a more complex nature. This is an important first step in the development of end to end automated scientific discovery systems in computational biology.

\begin{table}[htbp]
  \captionsetup[table]{singlelinecheck=off, justification=raggedright, width=0.8\linewidth}
  \caption{Percentage of successfull runs for each dataset-method combination. Method that failed produce any valid solution are assigned 0\%. Best per dataset are marked in bold.}
  \label{tab:max-metrics-sota}
  \begin{tabular}{lrrrrrr}
    \toprule
    \textbf{Method} & \textbf{AGO2} & \textbf{DE} & \textbf{HEC} & \textbf{HEE} & \textbf{NTP} & \textbf{OCRE} \\
    \midrule
    Agentomics-ML: gpt-4.1 (no feedback) & \textbf{100\%} & \textbf{100\%} & \textbf{100\%} & \textbf{100\%} & \textbf{60\%} & \textbf{100\%} \\
    Agentomics-ML: gpt4\_1 & \textbf{100\%} & \textbf{100\%} & \textbf{100\%} & \textbf{100\%} & \textbf{60\%} & \textbf{100\%} \\
    AIDE: gpt-4.1 & 0\% & 0\% & 0\% & 0\% & 40\% & 20\% \\
    DI: claude-3.7-sonnet & 0\% & 0\% & 20\% & 0\% & 20\% & 0\% \\
    DI: deepseek-chat & 0\% & 0\% & 0\% & 0\% & 20\% & 0\% \\
    DI: deepseek-r1 & 0\% & 0\% & 0\% & 0\% & 0\% & 0\% \\
    DI: llama-4-maverick & 0\% & 20\%& 0\% & 20\% & \textbf{60\%} & 0\% \\
    DI: gpt4.1 & 0\% & 0\% & 20\% & 40\% & 40\% & 0\% \\
    DI: qwen3-32b & 0\% & 0\% & 0\% & 0\% & 0\% & 0\% \\
    zero\_shot: claude-3.7-sonnet & 0\% & 20\% & 20\% & 20\% & 20\% & 0\% \\
    zero\_shot: deepseek-chat & 0\% & 20\% & 20\% & 40\% & 40\% & \textbf{100\%} \\
    zero\_shot: deepseek-r1 & 0\% & 40\% & 20\% & 40\% & 20\% & 20\% \\
    zero\_shot: llama-4-maverick & 0\% & 20\% & 0\% & 0\% & 0\% & 20\% \\
    zero\_shot: gpt-4.1 & 0\% & 20\% & 0\% & 0\% & 0\% & 0\% \\
    zero\_shot: o4-mini & 0\% & \textbf{100\%} & 40\% & 80\% & \textbf{60\%} & 60\% \\
    zero\_shot: qwen3-32b & 0\% & 0\% & 0\% & 20\% & 0\% & 20\% \\
    \bottomrule
  \end{tabular}
\end{table}

\subsection{Model performance}
We proceed to evaluate the performance of all benchmarked methods across datasets in Table \ref{tab:succ_rate}. We compare models produced from zero-shot LLMs and DI paired with several models,and AIDE and Agentomics-ML paired with gpt-4.1. We report the best run performance metric for each method on each dataset, as evaluated on a left-out test set (Table \ref{tab:succ_rate}). We also report the mean (Table \ref{tab:mean}) and standard deviation (Table \ref{tab:variance}) for the average performance metric over all runs. Missing values denote cases where the method did not produce any successful code over 5 tries. We also compare all methods against Human State of the Art (Human SOTA), those being published methodologies produced by domain experts, and we use metrics reported by those publications to allow for a fair comparison. All experiments were run using one NVIDIA A40 or A4000 GPU, 64 CPU cores and 32GB of RAM.

In the Drosophila\_enhancers\_stark dataset, Agentomics-ML outperforms all other methods (0.72 accuracy) as well as the Human SOTA from the original publication \citep{Greov2023} (0.59 accuracy). In the Human\_enhancers\_cohn dataset, Agentomics-ML without iterative feedback produces 0.72 accuracy, while the system using iterative feedback loop outperforms the other LLM-based methods with 0.74 accuracy. Similarly, for Human\_enhancers\_ensembl, Human\_nontata\_promoters, and Human\_ocr\_ensembl datasets, where the Agentomics-ML model with iterative feedback outperforms all LLM-based methods. In the AGO2\_CLASH\_Hejret dataset the iterative feedback loop improves the average precision score metric from 0.72 to 0.78. No other LLM-based methods managed to produce workable code for this dataset. We perform a t-test and find that the feedback loop statistically improves performance compared to non-feedback agent on 2 out of 6 datasets (\ref{tab:pvalue}). We find that models produced after iterative feedback obtain better test set metrics 80\% of the time and outperform no-feedback models relatively on-average by 3.8\%. This consistent improvement positions Agentomics-ML as the unanimous leader in automated machine learning on all tested genomic and transcriptomic datasets. It is possible that extending the number of iterations beyond what was tested, or adjusting the reflection mechanism, could push the benefits of the feedback loop even further.

\begin{table}[htbp]
  \captionsetup[table]{singlelinecheck=off, justification=raggedright, width=0.8\linewidth}
  \caption{Test set performance metrics for each dataset-method combination displaying maximum value of 5 separate runs. Accuracy is used for all except AGO2 where average precision score is used. Best overall is marked as underlined and bold. Best LLM-based method is marked in bold.}
  \label{tab:succ_rate}
  \begin{tabular}{lrrrrrr}
    \toprule
    \textbf{Method} & \textbf{AGO2} & \textbf{DE} & \textbf{HEC} & \textbf{HEE} & \textbf{NTP} & \textbf{OCRE} \\
    \midrule
    Agentomics-ML: gpt-4.1 (no feedback) & 0.724 & 0.716 & 0.716 & 0.864 & 0.897 & 0.786 \\
    Agentomics-ML: gpt4\_1 & \textbf{0.778} & \underline{\textbf{0.736}} & \textbf{0.743} & \textbf{0.885} & \textbf{0.925} & \textbf{0.816} \\
    AIDE: gpt-4.1 & N/A & N/A & N/A & N/A & 0.920 & 0.758 \\
    DI: claude-3.7-sonnet & N/A & N/A & 0.724 & N/A & 0.839 & N/A \\
    DI: deepseek-chat & N/A & N/A & N/A & N/A & 0.871 & N/A \\
    DI: deepseek-r1 & N/A & N/A & N/A & N/A & N/A & N/A \\
    DI: llama-4-maverick & N/A & 0.500 & N/A & 0.738 & 0.874 & N/A \\
    DI: gpt4.1 & N/A & N/A & 0.500 & 0.752 & 0.873 & N/A \\
    DI: qwen3-32b & N/A & N/A & N/A & N/A & N/A & N/A \\
    zero\_shot: claude-3.7-sonnet & N/A & 0.650 & 0.728 & 0.864 & 0.901 & N/A \\
    zero\_shot: deepseek-chat & N/A & 0.680 & 0.681 & 0.811 & 0.897 & 0.666 \\
    zero\_shot: deepseek-r1 & N/A & 0.700 & 0.621 & 0.568 & 0.874 & 0.458 \\
    zero\_shot: llama-4-maverick & N/A & 0.661 & N/A & N/A & N/A & 0.651 \\
    zero\_shot: gpt-4.1 & N/A & 0.703 & N/A & N/A & N/A & N/A \\
    zero\_shot: o4-mini & N/A & 0.708 & 0.728 & 0.848 & 0.885 & 0.506 \\
    zero\_shot: qwen3-32b & N/A & N/A & N/A & 0.850 & N/A & 0.662 \\
    \midrule
    Human SOTA & \underline{\textbf{0.860}} & 0.586 & \underline{\textbf{0.747}} & \underline{\textbf{0.933}} & \underline{\textbf{0.974}} & \underline{\textbf{0.825}} \\
    \bottomrule
  \end{tabular}
\end{table}

\section{Limitations} \label{sec:limitations}
\subsection{Potential pre-training data leakage }
As the training datasets of mainstream closed-source LLMs are not publicly disclosed by the providers, we cannot determine whether the datasets used for benchmarking in this study were part of the LLMs pre-training corpora. The same uncertainty applies to state-of-the-art approaches evaluated on these benchmarks, potentially raising concerns about data leakage and the integrity of comparative results.

\subsection{Reproducibility}
We used various available large language models for our experiments, including closed-source models that are available only through API calls. This brings a risk of a closed-source model being replaced or removed by the provider, limiting reproducibility of the results.  

\subsection{Datasets scope and generalizability}
Our evaluation is limited to a set of datasets from genomics and transcriptomics, selected for their accessibility and relevance. These datasets represent only a small subset of the computational biology landscape, and are solely focused on nucleotide sequence classification tasks. Computational biology ML tasks are much more varied, ranging from function or structure prediction, correlation of sequences and protein or transcript levels, value imputation, clustering, and many others. A future direction for this field is the exploration of different types of datasets and tasks. We believe that using our method offers better chances of generalizability, as seen in the AGO2\_CLASH\_Hejret dataset that had an atypical structure. Our approach had produced working code 100\% of the time, while no other approach managed to produce a single working model for this dataset.

\subsection{Prompt engineering and evaluation bias}
While we standardized user prompts to ensure fair comparisons, the iterative process of LLM-based development creates risks of unintentional overfitting. Iterative adjustments to hyperparameters, such as the number of loop iterations and temperature, or refinements of prompts based on observed performance can artificially inflate evaluation metrics. To mitigate this issue, we only used one dataset as a prototype (human\_nontata\_promoters), did not create per-dataset prompts, and used default hyperparameters where possible. All other datasets used the same prompt and hyperparameters that were never changed after testing. That said, it should be an important future task for an independent researcher to perform a benchmarking exercise in this field, ideally with never before published datasets.

\section{Conclusion} \label{sec:conclusion}
The advancement of high-throughput -omics technologies presents substantial analytical challenges due to data volume and complexity. Here, we introduce Agentomics-ML, a fully autonomous agent-based system that can automate machine learning experimentation for -omics classification tasks. A significant contribution is Agentomics-ML's high success rate (over 93\%) in generating executable machine learning code and trained models, notably on complex datasets, where other benchmarked agentic systems failed to produce working code. Our system demonstrates superior generalization and operational success compared to zero-shot LLM code generation and other evaluated agentic frameworks.

Agentomics-ML operates via a predefined, iterative workflow, autonomously executing sequential stages including data exploration, representation, model architecture design, training, and inference script generation, primarily utilizing a bash shell interface for file system interaction and tools to write and execute python code. A key feature is its reflection loop, where training and validation metrics inform subsequent iterations, facilitating model refinement and ensuring the generation of reproducible artifacts. While state-of-the-art models developed by domain experts currently achieve superior absolute performance in most datasets, Agentomics-ML substantially reduces this disparity for fully autonomous systems and outperforms human state-of-the-art on one dataset. Moreover, the average cost of one agent run is less than 2\$.
Future work will focus on extending Agentomics-ML's applicability to a broader array of biological data modalities and computational tasks. Agentomics-ML underscores the considerable potential of autonomous agent-based systems to advance data-driven discovery in molecular biology.

\bibliographystyle{plainnat}
\bibliography{references}

\appendix
\newpage
\section{Supplementary Tables} \label{app:tables}
\begin{table}[!ht]
\caption{Test set performance metrics for each dataset-method combination displaying mean
value of 5 separate runs. Accuracy is used for all except AGO2 where average precision score is
used. Best overall is marked as underlined and bold. Best LLM-based method is marked in bold.}
\label{tab:mean}
    \centering
    \begin{tabular}{lrrrrrr}
    \toprule
    \textbf{Method} & \textbf{AGO2} & \textbf{DE} & \textbf{HEC} & \textbf{HEE} & \textbf{NTP} & \textbf{OCRE} \\
    \midrule
        Agentomics: gpt4\_1 (no feedback) & 0.698 & 0.673 & 0.698 & 0.796 & 0.881 & 0.758 \\ 
        Agentomics: gpt4\_1 & \textbf{0.726} & \textbf{0.715} & \textbf{0.729} & 0.827 & 0.899 & \textbf{0.787} \\ 
        AIDE: gpt-4.1-2025-04-14 & N/A & N/A & N/A & N/A & \textbf{0.905} & 0.757 \\ 
        DI: claude-3.7-sonnet & N/A & N/A & 0.724 & N/A & 0.839 & N/A \\ 
        DI: deepseek-chat & N/A & N/A & N/A & N/A & 0.871 & N/A \\ 
        DI: deepseek-r1 & N/A & N/A & N/A & N/A & N/A & N/A \\ 
        DI: llama-4-maverick & N/A & 0.5 & N/A & 0.738 & 0.801 & N/A \\ 
        DI: gpt-4.1-2025-04-14 & N/A & N/A & 0.5 & 0.706 & 0.872 & N/A \\ 
        DI: qwen3-32b & N/A & N/A & N/A & N/A & N/A & N/A \\ 
        zero\_shot: claude-3.7-sonnet & N/A & 0.650 & 0.728 & \textbf{0.864} & 0.901 & N/A \\ 
        zero\_shot: deepseek-chat & N/A & 0.680 & 0.681 & 0.797 & 0.882 & 0.639 \\ 
        zero\_shot: deepseek-r1 & N/A & 0.700 & 0.621 & 0.765 & 0.874 & 0.458 \\ 
        zero\_shot: llama-4-maverick & N/A & 0.661 & N/A & N/A & N/A & 0.651 \\ 
        zero\_shot: gpt-4.1-2025-04-14 & N/A & 0.703 & N/A & N/A & N/A & N/A \\ 
        zero\_shot: o4-mini & N/A & 0.694 & 0.719 & 0.800 & 0.868 & 0.609 \\ 
        zero\_shot: qwen3-32b & N/A & N/A & N/A & 0.850 & N/A & 0.662 \\ 
        \bottomrule
    \end{tabular}
\end{table}

\begin{table}[!ht]
\caption{P-values for a t-test comparing Agentomics-ML with and without feedback mechanism.}
\label{tab:pvalue}
    \centering
    \begin{tabular}{lr}
    \toprule
        \textbf{Dataset} & \textbf{p\_value} \\
        \midrule
        human\_ocr\_ensembl & 0.022 \\ 
        AGO2\_CLASH\_Hejret2023 & 0.356 \\ 
        human\_nontata\_promoters & 0.187 \\ 
        human\_enhancers\_cohn & 0.011 \\ 
        drosophila\_enhancers\_stark & 0.068 \\ 
        human\_enhancers\_ensembl & 0.200 \\ 
        \bottomrule
    \end{tabular}
\end{table}

\begin{table}[!ht]
\caption{Test set performance metrics for each dataset-method combination displaying variance between the 
5 separate runs. Accuracy is used for all except AGO2 where average precision score is used.}
\label{tab:variance}
    \centering
    \begin{tabular}{lrrrrrr}
    \toprule
    \textbf{Method} & \textbf{AGO2} & \textbf{DE} & \textbf{HEC} & \textbf{HEE} & \textbf{NTP} & \textbf{OCRE} \\
    \midrule
    
        Agentomics: gpt4\_1\_no\_feedback & 0.030 & 0.036 & 0.030 & 0.039 & 0.014 & 0.019 \\ 
        Agentomics: gpt\_1 & 0.033 & 0.019 & 0.016 & 0.045 & 0.023 & 0.023 \\ 
        AIDE: gpt-4.1 & N/A & N/A & N/A & N/A & 0.021 & N/A \\ 
        DI: claude-3.7-sonnet & N/A & N/A & N/A & N/A & N/A & N/A \\ 
        DI: deepseek-chat & N/A & N/A & N/A & N/A & N/A & N/A \\ 
        DI: deepseek-r1 & N/A & N/A & N/A & N/A & N/A & N/A \\ 
        DI: llama-4-maverick & N/A & N/A & N/A & N/A & 0.103 & N/A \\ 
        DI: gpt-4.1-2025-04-14 & N/A & N/A & N/A & 0.065 & 0.001 & N/A \\ 
        DI: qwen3-32b & N/A & N/A & N/A & N/A & N/A & N/A \\ 
        zero\_shot claude-3.7-sonnet & N/A & N/A & N/A & N/A & N/A & N/A \\ 
        zero\_shot: deepseek-chat & N/A & N/A & N/A & 0.020 & 0.022 & 0.042 \\ 
        zero\_shot: deepseek-r1 & N/A & 0.000 & N/A & 0.004 & N/A & N/A \\ 
        zero\_shot: llama-4-maverick & N/A & N/A & N/A & N/A & N/A & N/A \\ 
        zero\_shot: gpt-4.1-2025-04-14 & N/A & N/A & N/A & N/A & N/A & N/A \\ 
        zero\_shot: o4-mini & N/A & 0.014 & 0.014 & 0.032 & 0.019 & 0.107 \\ 
        zero\_shot: qwen3-32b & N/A & N/A & N/A & N/A & N/A & N/A \\ 
    \bottomrule   
    \end{tabular}
\end{table}

\end{document}